\documentclass{article}
\usepackage{spconf,amsmath,graphicx}
\usepackage{subfigure}
\usepackage{makecell}
\usepackage{comment}
\usepackage{diagbox}

\title{Multistream ValidNet: Improving 6D Object Pose Estimation by Automatic Multistream Validation}
\name{\protect\parbox{\textwidth}{\protect\centering Joy Mazumder, Mohsen Zand, and Michael Greenspan \\ \{joy.mazumder, 
m.zand,  michael.greenspan\}@queensu.ca \thanks{Thanks to Bluewrist Inc. and NSERC for their support of this work.}}}
\address{Dept. Electrical and Computer Engineering, Ingenuity Labs, Queen's University, Kingston, Canada}

\begin{document}
\nocite{mazumder2020validnet} 
%

\onecolumn
    $\copyright$ 2021 IEEE. Personal use of this material is permitted. Permission from IEEE must be obtained for all other uses, in any current or future media, including reprinting/republishing this material for advertising or promotional purposes, creating new collective works, for resale or redistribution to servers or lists, or reuse of any copyrighted component of this work in other works.
\twocolumn

\maketitle

\begin{abstract}
This work 
presents a novel approach
to improve the 
results of
pose estimation 
by detecting and distinguishing
between the occurrence
of True and False Positive results. 
It achieves this by
training a binary classifier on the 
output of an arbitrary pose estimation algorithm,  
and returns a binary label indicating the validity of the result. We demonstrate that our approach improves upon 
a state-of-the-art pose estimation result on the Siléane dataset,
outperforming 
a variation of the alternative CullNet method by 4.15\% in average class accuracy and 0.73\% in overall accuracy at validation.
Applying our method can also improve the pose estimation average precision results 
of Op-Net by 6.06\% on average.

\end{abstract}

\section{Introduction}
\label{sec:intro}

Object pose estimation deals with detecting objects and estimating their orientations and translations with respect to a canonical basis. It is a crucial task which can assist a variety of applications in different areas such as robotic manipulation, scene understanding, autonomous driving, and augmented reality. In many applications such as robotic bin picking, accurate pose estimation is essential since a false positive fault can lead to a catastrophic system damage and failure. Pose estimation algorithms therefore need to achieve nearly perfect results. Current algorithms are however far from success, and any effort for their improvement is worth exploring. 

\begin{figure}[t]
\begin{center}
   \subfigure[Pose estimation using OP-Net~\cite{kleeberger2020single} with confidence score$>$0.5, with True Positive (TP, green) and False Positive (FP, red) overlays.]{\includegraphics[width=\linewidth]{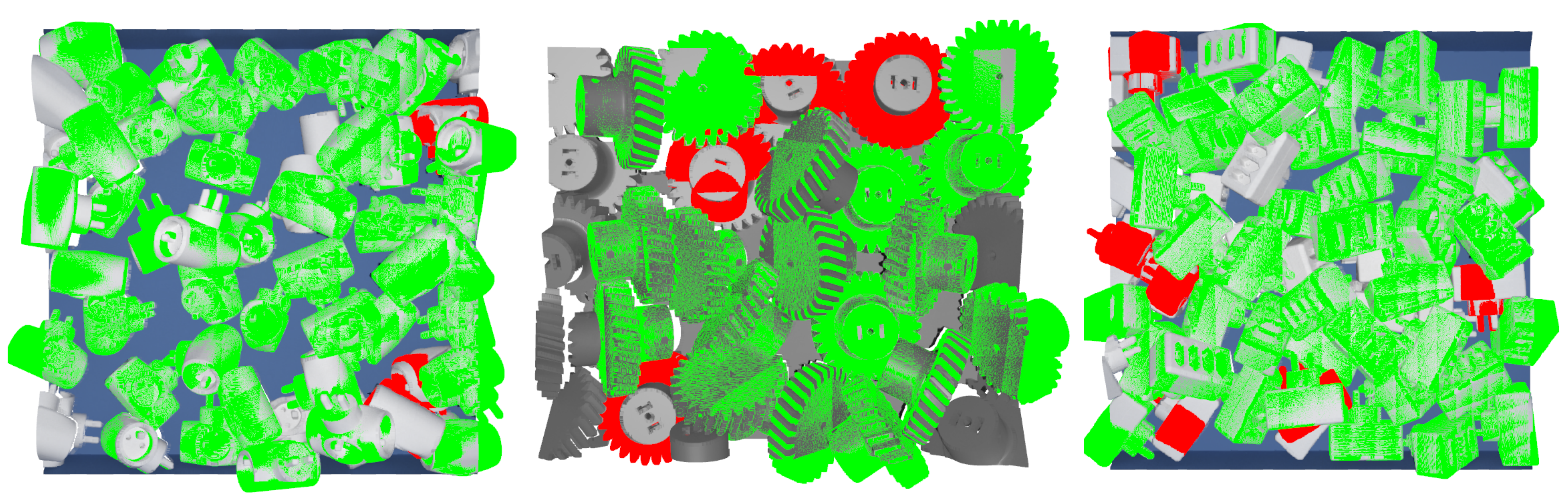}}%
   
   \subfigure[FP poses from (a) detected with Multistream Validnet and removed.]{\includegraphics[width=\linewidth]{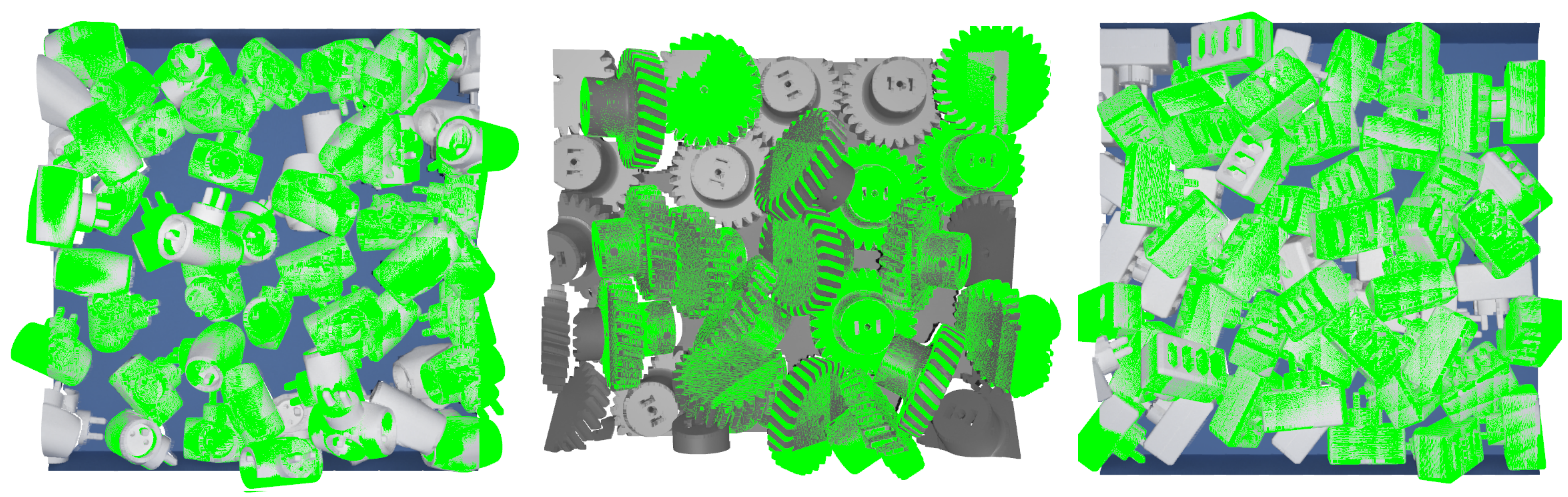}}%
  
    \caption{Pose estimation before and after Multistream ValidNet: green is True Positive (TP), red is False Positive (FP).}
    \vspace{-0.75 cm}
    \label{fig:6Dof_results}
\end{center}
\end{figure}

Recently, it has been shown that current state-of-the-art algorithms can be further improved by detecting and correcting their false detections~\cite{gupta2019cullnet, mazumder2020validnet}. In~\cite{gupta2019cullnet}, for example, Gupta et al. investigated the problem of inaccurate confidence values which are used by many pose estimation approaches for final object pose prediction. They proposed CullNet to calibrate the confidence scores of the pose proposals. Mazumder et al.~\cite{mazumder2020validnet} proposed ValidNet which used a two-class classifier to discriminate true positive and false positive instances in 3D surface registration algorithms. OP-Net~\cite{kleeberger2020single} proposed a 3 dof confidence score (only for the translational pose component) for differentiating true and false object detections. Generally, there are two main strategies which can be used to improve the pose estimation results. 1. False positive (FP) detections can be detected and removed, thereby reducing FPs and increasing true negatives (TN). 2. FPs can be identified and further processed, thereby converting some of them into true positive (TP), and some into TNs. Thus the FP rate is reduced, and both the TP and TN rate is increased.

In this paper, we propose to use two different streams within a network for differentiating correct and incorrect poses. One stream of our network operates on depth images and another stream of our network operates on point cloud stream. Both streams predict the correct and incorrect alignment independently and finally we are taking the scoring average of the both steams which can outperform any single stream and other proposed methods for validation.   

Our proposed method is a post-processing algorithm which is applied to the output of an initial pose determination phase. Particularly, the correctness of the estimated 6 dof poses of a known class are returned at this phase. In addition to the 6D pose estimates, the multistream input to our method comprises both a depth stream and a point cloud stream of the scene, as well as a 3D model of the known object class. The output is the binary classification of each of the input pose estimates as either true (\emph{valid}) or false (\emph{invalid}).

\begin{figure*}[t]
    \centering
    \includegraphics[width=0.7\linewidth]{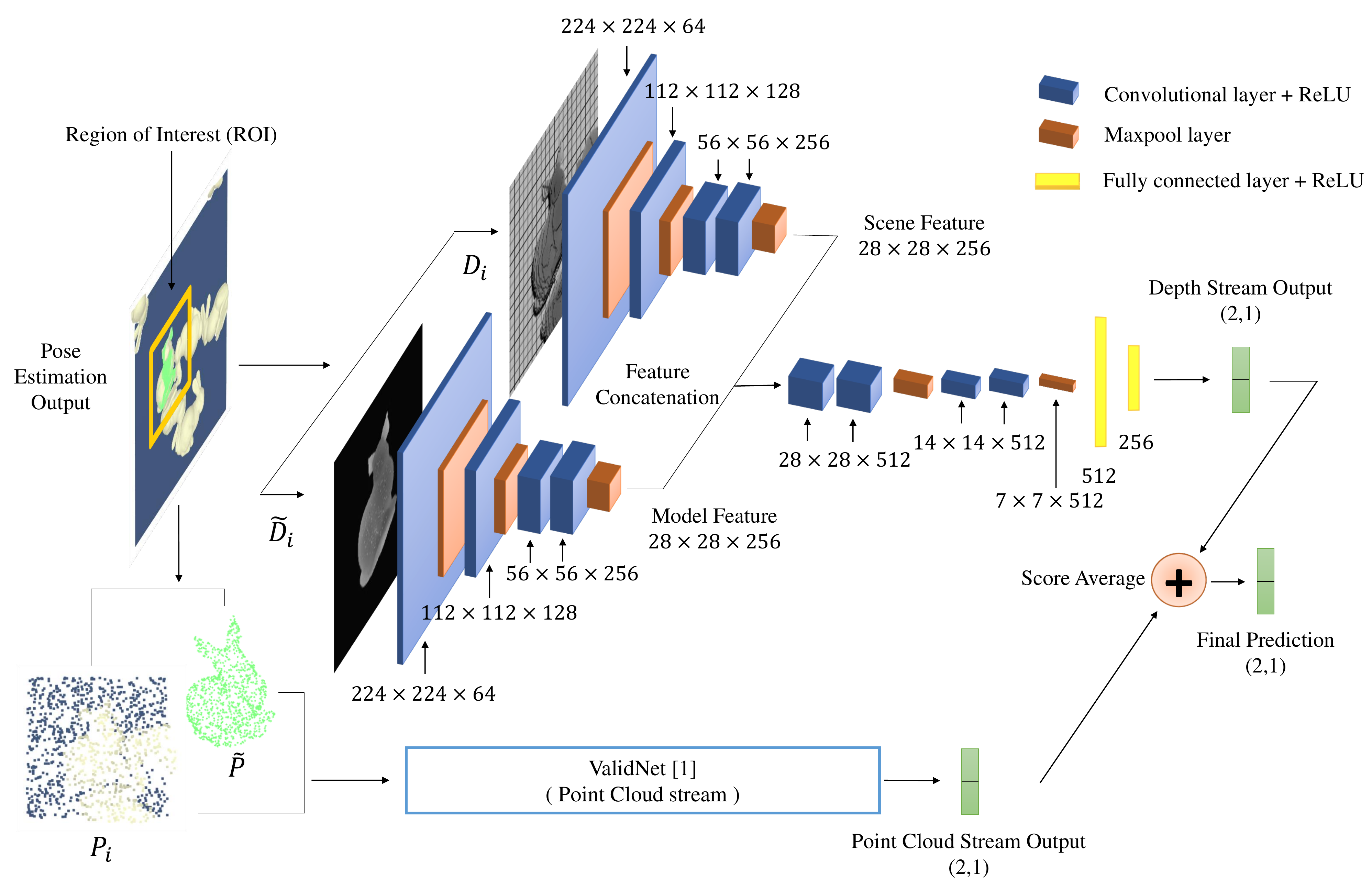} 
    \caption{Multistream ValidNet architecture, with depth stream shown in detail, and point cloud stream~\cite{mazumder2020validnet} encapsulated. }
    \label{fig:architecture}
\end{figure*}

\section{Related Work}

Object pose estimation is a well-studied research problem and an active research direction in computer vision. Early approaches were based on aligning 3D CAD models to 3D point clouds using  hand-crafted features~\cite{drost2010model,rusu2009fast} and variants of iterative closest point algorithm~\cite{besl1992method}, using either depth or RGB-D sensors~\cite{zeng20173dmatch}.  
Recently, CNN-based deep learning algorithms have been developed to estimate the 6 dof pose of an object in a single shot ~\cite{kleeberger2020single, dong2019ppr, sock2018multi,li2018unified}.  For example, PoseNet~\cite{kendall2015posenet} proposed a CNN model to regress a 6 dof camera pose from a single RGB image,
whereas PoseCNN~\cite{xiang2017posecnn} localized objects in the 2D image and predicted their depth information to generate 3D locations.

Another recent line of research has focused on pose refinement by improving or validating the output of pose estimation algorithms. 
In Bariya and Nishino~\cite{bariya2010scale}, a threshold set was used on the number of supporting correspondences and a collection of hypotheses was achieved for each model. 
Papzov and Burschka~\cite{papazov2010efficient} introduced an acceptance function to prune hypotheses and validate the model-scene correspondences. Aldoma et al.~\cite{aldoma2012global} investigated the entire set of hypotheses as a global scene model and minimized a global cost function. Gupta et al.~\cite{gupta2019cullnet} addressed the problem of unreliable object pose confidences produced by CNNs. They found false positives from different object pose proposals and performed a refinement algorithm by calibrating the confidence scores using knowledge of pose hypotheses.

\section{Proposed Multistream ValidNet}
\label{sec:proposed_method}

The goal of Multistream ValidNet is to  classify the output of a pose estimator as either 
\emph{valid} or 
\emph{invalid}. 
A pose estimator takes as input a depth image or point cloud $I$ containing instances of a known object, 
and returns estimates 
$\{\hat{\theta_i}\}_1^k$ 
of
the $k$ object instances' 6 dof poses.
Multistream ValidNet in turn takes as input
$I$ and 
$\{\hat{\theta_i}\}_1^k$,
and returns binary labels
$\{B_i\}_1^k$,
indicating whether the pose estimator respectively returned
a \emph{valid} (TP) 
or \emph{invalid} (FP) result
for each 
$\hat{\theta_i}$.


To classify the 
$\{\hat{\theta_i}\}_1^k$,
we compose from $I$ and each $\hat{\theta_i}$ a pair of two depth images ($D_i$ and $\tilde{D}_i$) 
and two point clouds
($P_i$ and $\tilde{P}$) to serve as input to the respective depth and point cloud streams of the network. 
Here, $D_i$ and $P_i$ are cropped and sampled from $I$,
whereas $\tilde{D}_i$
are generated synthetically from the 
estimated pose $\hat{\theta_i}$
and the 
3D object model and $\tilde{P}$ is the model point cloud.

For $D_i$ and $\tilde{D}_i$ composition, we first transform 
the 3D model of the known object class by $\hat{\theta_i}$ 
into an initially empty scene. We then create a synthetic depth image of the transformed model within the scene, and crop from it a
region of interest (ROI$_D$) to form
$\tilde{D}_i$.
ROI$_D$ is centered at 
$\hat{\theta_i}$
and its extent is based on the model dimension ($\pm 1.2\times$ object radius from detected model centroid). 
This same ROI$_D$ is then used to crop the corresponding
$D_i$ from $I$.
and both 
$\tilde{D}_i$
and $D_i$
are then fed into the depth stream of the proposed network. 

For the point cloud stream, 
the model point cloud
$\tilde{P}$ represents the model canonical pose,
and therefore need only be generated once for all
values of $\theta_i$.
A region of interest ROI$_P$ is generated from the inverse of the transformation matrix
that represents
$\hat{\theta_i}$,
as in~\cite{mazumder2020deep,mazumder2020validnet}.
Applying the inverse transformation
relieves the network of the need to adapt to different translations and rotations 
of the model in the scene, 
and it is therefore able to learn strong discriminative characteristics of the
single canonical model pose.
The
point cloud is then cropped
with  ROI$_P$ to form 
$P_i$.
The transformed point clouds
$P_i$
of the scene and 
$\tilde{P}$
of the model are then fed as input into the point cloud stream.

The architecture of Multistream ValidNet is shown in Fig.~\ref{fig:architecture}, in which the novel (top) depth stream is illustrated in detail, 
and the (bottom) point cloud stream which has been previously reported~\cite{mazumder2020validnet} is encapsulated. In the depth stream, we initially used two identical but differently weighted blocks (consisting of four $3\times 3$ convolutional layers and three $2\times 2$ maxpool layers)  for extracting features from the model and the scene independently. These features are able to characterize the model and segment orientations.  

Following feature extraction, the next step is to determine the similarity of the two
feature sets, and consequently the validity of the detected object pose. We combine the feature sets by concatenating them along the feature axis as shown in Fig.~\ref{fig:architecture}. Following concatenation, we apply a combined feature extractor block (consisting of
a pair of 2 convolutional layers followed by a maxpool layer) which accumulates the features required to find the validity of pose determination. Finally, 2 fully connected layers are used to 
binary classify the valid and invalid poses, based on the extracted combined features of the network.

 For the MultiStream ValidNet point cloud stream, we have used our previous approach ~\cite{mazumder2020validnet,mazumder2020deep}. For this stream, we have used the PointNet~\cite{qi2017pointnet} segmentation network to extract the features from the model and cropped scene. These features are used to find the point wise correlation between the model point and segment points, using a point wise dot product of each model point's features with each segment point's features. After the correlation, max-pool is applied to capture the maximum feature match for all the model points. Finally, fully connected layers are used to classify the correct and incorrect detections, which is output as a fully connected binary layer. To combine both streams, we have fused the prediction of the depth image stream and point cloud stream by taking the score average of their predictions.








\section{Experiments}

\subsection{Object pose estimation}
Since our proposed method accepts object poses as input, we need a pose estimation network to train and evaluate the effectiveness of our method. For the estimation of 6 dof object pose from the scene, we have used OP-Net~\cite{kleeberger2020single} proposed by Kleeberger and Huber. OP-Net was developed to address the problem of pose determination for the robotic bin-picking application, where a scene comprises multiple instances of a single known object class, as shown in Fig.~\ref{fig:6Dof_results}(a). OP-Net takes as input a depth image of the scene and predicts the 6 dof poses of the multiple known object instances. 

The network divides the input scene into $S\!\times\!S$ volume elements and predicts one object pose per element. 
When there are multiple object centers in a single volume element, it returns the pose of the object that is most visible. Each volume element generates an 8D vector as output, comprising 3 translation and 3 rotation dimensions, a visibility score and a confidence score for the object center lying in that volumetric element. For training OP-Net author proposed two different loss function. Here, we used $\mathcal{L}_{ori1}$ loss as it was providing better results for most of the objects in the Sil\'{e}ane dataset~\cite{kleeberger2020single}.

\setlength{\tabcolsep}{4pt}
{\renewcommand{\arraystretch}{1.5}
\begin{table*}[t]
	\centering
	\scriptsize
	\begin{tabular}{c c c c c c c c c cc cc cc cc| c c }
		  \hline
		\multicolumn{1}{c}{} & \multicolumn{2}{c}{\textbf{Bunny}} 
		& \multicolumn{2}{c}{\textbf{TLess22}} 
		& \multicolumn{2}{c}{\textbf{Pepper}}
		& \multicolumn{2}{c}{\textbf{Gear}}
		& \multicolumn{2}{c}{\textbf{CandleStick}}
		 & \multicolumn{2}{c}{\textbf{Brick}}
		  & \multicolumn{2}{c}{\textbf{TLess20}}
		   & \multicolumn{2}{c|}{\textbf{TLess29}}
		    & \multicolumn{2}{c}{\textbf{Average}}
		     \\ \cline{2-19} 
		\multicolumn{1}{c}{} & ACA              & OA               & ACA               & OA               & ACA               & OA 
		& ACA               & OA
		& ACA               & OA
		& ACA               & OA
		& ACA               & OA
		& ACA               & OA
			& ACA               & OA
		\\ \hline \hline 
		
		CullNet~\cite{gupta2019cullnet}                & 
		93.63 & \textbf{99.52} & 93.73 & 95.56 & 84.17 & 99.43 & 98.75 & 98.85 & 77.39 & \textbf{96.21} & 96.45 & 96.61 & 95.19 & 98.15 & 90.78 & 95.15 & 91.26 & 97.44

		\\ 
		\hline
		 \makecell[c]{ Multistream ValidNet\\ Depth stream  }                &     94.58 & 99.54 & 95.91 & 97.35 & 89.87 & 99.53 & 99.41 & 99.5 & 89.44 & 93.24 & 97.58 & 97.96 & 96.7 & 98.72 & 92.33 & 97.31 & 94.48 &	97.89

		\\ \hline
		
		\makecell[c]{ValidNet~\cite{mazumder2020validnet}  \\(Point Cloud stream)}              & 95.31 & 99.13 & 96.61 & 97.75 & 92.49 & 99.14 & \textbf{99.52} & \textbf{99.7} & 87.97 & 94.59 & 96.93 & 97.18 & 96.87 & 98.29 & 92.63 & 97.02 & 94.79 &	97.85

		\\ \hline
		Multistream ValidNet           &     \textbf{95.49} & 99.48 & \textbf{96.64} & \textbf{97.82} & \textbf{92.69} & \textbf{99.53} & \textbf{99.52} & \textbf{99.7} & \textbf{90.66} & 94.55 & \textbf{97.68} & \textbf{98.01} & \textbf{96.95} & \textbf{98.63} & \textbf{93.66} & \textbf{97.67} & \textbf{95.41}	& \textbf{98.17}
		
		\\ \hline
		
	\end{tabular}
	\caption{Comparison of Average Class Accuracy (ACA \%)
	and 
	Overall Accuracy (OA \%) for
	Different
	Architectures}
	\label{tab:validationresults}
\end{table*}
}

\setlength{\tabcolsep}{4pt}
{\renewcommand{\arraystretch}{1.5}
\begin{table*}[ht]
	\centering
	\scriptsize
	\begin{tabular}{c c c c c c c c c| c}
		\hline
		\multicolumn{1}{c}{} & 
		\multicolumn{1}{c}{\textbf{Bunny}} &
		\multicolumn{1}{c}{\textbf{TLess22}} &
		\multicolumn{1}{c}{\textbf{Pepper}} &	
		\multicolumn{1}{c}{\textbf{Gear}} &		\multicolumn{1}{c}{\textbf{CandleStick}} &
		\multicolumn{1}{c}{\textbf{Brick}} &
		\multicolumn{1}{c}{\textbf{TLess20}} &
		\multicolumn{1}{c|}{\textbf{TLess29}} &
		\multicolumn{1}{c}{\textbf{Average}}
		
		\\ \hline  \hline 
		
		OP-Net + PP ~\cite{kleeberger2020single}                & 97.51\% & 94.57\% & 97.44\% & 83.84\% & 97.41\% & 52.34\% & 93.8\% & 72.48\%  & 86.17\%

		\\ 
		\hline
		OP-Net + PP +  Multistream ValidNet                 &     \textbf{98.61\%} & \textbf{96.97\%} & \textbf{98.61\%} & \textbf{91.18\%} & \textbf{98.61\%} & \textbf{73.04\%} & \textbf{96.12\%} & \textbf{84.66\%} & \textbf{92.23\%}
		
		\\ \hline

	\end{tabular}
	\caption{Performance Comparison of Pose Estimation Before and After Applying Multistream ValidNet }
	\label{tab:poseresults}
\end{table*}


\subsection{Dataset}
We have trained
both OP-Net~\cite{kleeberger2020single}  and Multistream ValidNet on  
8 objects from the
Fraunhofer IPA dataset~\cite{kleeberger2019large}.
We then 
 tested using these same 8 objects
as collected in the  
Sil\'{e}ane ideal  dataset~\cite{bregier2017iccv}, 
which is similar to the training and testing protocol followed by OP-Net. 

To train Multistream ValidNet, we first trained OP-Net using relatively few epochs,
between 2 and 10 depending on the particular object. The reason for this was that we required a 
balance of both TP and FP pose estimates to facilitate training,
and so early termination kept the
accuracy of OP-Net at approximately $50\%$.
Rather than training Multistream ValidNet on the entire IPA dataset, we only used a subset to accelerate the training time to a feasible range. For defining TP and FP pose estimates, we used the metric proposed by Br\'{e}gier et al.~\cite{bregier2018defining} which consider the object's rotational and cyclic symmetry. Depending on the symmetry of the object, the authors proposed to represent the pose of an object (rotation and translation) by a set of high dimensional points $\mathbf{p}\in \mathcal{R}(\theta)$. The distance between the detected pose $\hat{\theta_i}$ and the ground truth pose $\theta_{gt}$ of a particular object is defined by the minimum Euclidean distance between their respective pose representatives:
\begin{equation}
    d(\hat{\theta_i},\theta_{gt}) = \mbox{min} \Vert \mathbf{p_i}-\mathbf{p_{gt}} \Vert
    \;\; \forall \;\; 
    \mathbf{p_i}\!\in\!\mathcal{R}(\hat{\theta_i}), \mathbf{p_{gt}}\!\in\!\mathcal{R}(\theta_{gt})
\end{equation}

A detected object pose, $\hat{\theta_i}$ will be define as TP, if the pose distance $ d(\theta_i,\theta_{gt})$ is less than 10\% of the object's diameter.

Using this definition, we have partitioned the results into the positive and the negative class sets. We used 50000 positive classes and 50000 negative classes per object. We only used predictions for which the OP-Net confidence score was greater than 0.3, 
as anything less than that was 
more likely to yield an empty cell which is of little value in our training. For testing our network, we first trained OP-Net for each object
of the Sil\'{e}ane Dataset
using 50 epochs, so that it
would converge fully and
result in a higher probability of
TP during inference. 

After pose prediction, we transformed the model into the scene and created $D_i$ and $\tilde{D}_i$ for the depth stream of the network. Our depth stream takes input images of size $224\times 224$. To reduce the overfitting problem, we did some data augmentation during training. We initially resized the images into $236\times 236$ and then took some random identical crop on both images of size $224\times 224$. We also randomly rotated both images by increments of $90^o$. For the point cloud stream, we generated $P_i$ and $\tilde{P}$, each of size 1024 points. We randomly shuffled the $P_i$ point in order to reduce overfitting.

\subsection{Setup and Environment}
We trained both streams of our network independently and averaged their output scores. To train both streams of the network, we used Adam optimizer with a softmax cross-entropy loss function. The initial learning rate was 0.0001 for both streams, which was then divided by 10 every 10 epochs. The network was trained for 60 epochs with batch size of 8. The parameters of the point cloud stream were selected exactly as in our previous work~\cite{mazumder2020validnet}. For the depth stream of the network, the parameters of the convolutional blocks were selected based on VGG-11 ~\cite{simonyan2014very} and the parameters of the fully connected layers were selected using grid search. We used ReLU activation function in all the layers and used drop out with keep probability of 0.7 in the last fully connected hidden layer. All the networks are trained independently per object.

\subsection{Results}

We have compared Multistream ValidNet against the two purely single streams of the network,
as well as a variation of CullNet~\cite{gupta2019cullnet}, which was originally designed for calculating pose aware confidence scores of 6 dof poses from RGB images. To validate the TP and FP results, 
we have modified CullNet in two ways,
the first being rather than
RGB images,  we feed it with  the depth images
$D_i$ and $\tilde{D}_i$,
the same as in the Multistream ValidNet depth stream. 
The second change was, instead of 
a regression output
as applied  in the original
CullNet work, 
we have applied a classification output.
To achieve this, we replaced the 
final regression output layer with a 
softmax binary classification layer.
The results of classifying TP and FP detection are shown in Table~\ref{tab:validationresults}, which shows both average class accuracy (ACA) and overall accuracy (OA). 
From the table, it is clear that Multistream ValidNet  outperforms Cullnet method by 4.15\% in ACA and 0.73\% in OA. It also outperforms any single stream of our network to differentiate TP and FP results on the Sil\'{e}ane dataset.

In the Sil\'{e}ane dataset, the authors proposed using average precision to measure pose estimation performance. To demonstrate the effectiveness of our proposed method, we took the recent state-of-the-art pose prediction method OP-Net and replaced their predicted confidence score by Multistream ValidNet's TP probability. The OP-Net confidence score is only for the 3 dof translational component. This confidence score can be high if the translational components are correct but the full 6 dof predicted poses are incorrect.   Since our proposed method has a pose aware confidence score replacing OP-Net's confidence score,  our method outperforms OP-Net by 1.1\%-20\% depending on the object. The detailed results for all objects are provided in Table~\ref{tab:poseresults}.

\section{Conclusion}
 Differentiating between True and False Positive results can be beneficial in any pose estimation task, and is particularly crucial for robotic manipulation tasks. To achieve this goal, 
we have proposed a multistream architecture which uses depth images and point clouds of the model and the scene to distinguish 
 between TP and FP results which outperform a variation of CullNet which was designed to solve a similar problem. In future work, we have plan to investigate the RGB stream in some other datasets where the objects in the scene have color variation. We will also explore the options of further refining the estimated poses when validation detects invalid poses,
 potentially to improve the 6 dof pose estimation accuracy, thereby allowing the conversion of FP to TP results.

\bibliographystyle{IEEEbib}
\bibliography{refs}

\end{document}